# Sentiment Classification in Bangla Textual Content: A Comparative Study

Md. Arid Hasan,[1] Jannatul Tajrin,[1] Shammur Absar Chowdhury,[2] Firoj Alam[2]
[1]Cognitive Insight Limited, Bangladesh
[2]Qatar Computing Research Institute, HBKU, Qatar
[1]{arid.hasan.h, jannatultajrin33}@gmail.com, [2]{fialam, shchowdhury}@hbku.edu.qa

*Abstract*—Sentiment analysis has been widely used to understand our views on social and political agendas or user experiences over a product. It is one of the cores and well-researched areas in NLP. However, for low-resource languages, like Bangla, one of the prominent challenge is the lack of resources. Another important limitation, in the current literature for Bangla, is the absence of comparable results due to the lack of a well-defined train/test split. In this study, we explore several publicly available sentiment labeled datasets and designed classifiers using both classical and deep learning algorithms. In our study, the classical algorithms include SVM and Random Forest, and deep learning algorithms include CNN, FastText, and transformer-based models. We compare these models in terms of model performance and time-resource complexity. Our finding suggests transformer-based models, which have not been explored earlier for Bangla, outperform all other models. Furthermore, we created a weighted list of lexicon content based on the valence score per class. We then analyzed the content for high significance entries per class, in the datasets. For reproducibility, we make publicly available data splits and the ranked lexicon list. The presented results can be used for future studies as a benchmark.

*Index Terms*—Sentiment Analysis, SVM, Random Forest, Fast-Text, CNN, Transformers models

## I. Introduction

The wide use of social media sites and other digital technologies are now integrated with our daily life. Every day we share information, voice our opinions and concerns through these platforms. Access to such enormous and opinionated digital content gives the research community an opportunity to analyze and summarize the content in order to help in a decision making process, understand the sentiment of a product, or views on cultural, social, international, and political agendas [1]. It is one of the well studied and active research areas in NLP. The research mostly focused on classifying sentiment in either one of two labels (i.e., positive or negative) or five labels (i.e., very positive to very negative) from textual information such as movie-reviews [2], newspaper articles and comments [3], and emotion from spoken conversations [4].

The majority of the existing research is limited to English and other resource-rich languages, due to the availability of large public resources and benchmarks. Earlier, one of the commonly used approaches is to use sentiment lexicon (e.g., SentiWordNet, Sentiment Treebank, and Psycholinguistic features) [5], [6] as features for designing the sentiment classifier. Considering it as a text classification problem, it has been widely addressed using machine learning – from classical to deep learning algorithms. Unlike these resource-rich languages, studies and resources for Bangla are still very limited. This limitation is the results of *(i)* having a very few publicly available datasets and *(ii)* inconsistent benchmarks due to incompatible train/dev/test splits (except in 1-2 datasets). From the modeling perspective, the current literature explored both classical (e.g., SVM, Random Forest, Naive Bayes) and deep learning algorithms (e.g., Convolution Neural Network (CNN), Long Short-Term Memory (LSTM)). The literature lacks the use of the recent successful algorithms such as Transformer models [7].

To address such limitations in this study we conduct comparative experiments by utilizing publicly available Bangla sentiment datasets, and our in house developed dataset. We experimented with both classical and deep learning algorithms in order to understand the performance difference – in terms of training complexity, model performance, and implication in real deployment (e.g., deep learning algorithms requires GPU to train and for inference). The motivation is to understand the trade-off between computational demand and performance. While doing so, our contributions of this study are as follows:

- We curated publicly available datasets, carefully cleaned and split into the train, dev, and test set. We make them available for reproducibility.[1]
- We analyze the datasets to better understand the lexicons that significantly reflect the positive and negative sentiment.
- We experiment with both classical, and deep learning algorithms, that are widely used – Random Forest, SVM, CNNs, Transformer models among others; and report our findings for future reference.

## II. Related Work

The rise of social media has increased the interest among the researchers to analyze the sentiment using social media data. The rapid advancement of sentiment analysis started back in the early 2000s. Earlier studies include document [8], [9], sentence and aspect level [10] classification tasks.

---

[1]https://github.com/banglanlp/bangla-sentiment-classification

The initial study of sentiment analysis for Bangla language started back in the 2010s [11]. In [11], authors developed a corpus-based WordNet for sentiment analysis named "SentiWordNet", in which authors claimed that uses of SentiWordNet are reliable. In another study [12], authors reported an SVM-based polarity classification task on news data, in which authors achieved a precision 70.04% and a recall 63.02%. However, the authors also used Part-of-Speech (PoS) tagging, SentiWordNet, stemming, and negative word list. The study in [13] proposed a semi-supervised technique for Bangla sentiment analysis using Twitter posts and achieve an accuracy of 93% using unigrams with emoticons. The authors also used preprocessing steps such as tokenization, normalization, and PoS tagging before classifying the data. For classifying, the authors used NLTK toolkit for designing the SVM and the MaxEnt model. However, the dataset is not publicly available and the results are not reproducible.

In the study of [14], authors report SVM models in combination with N-gram for sentiment analysis on Bangla news comments, in which authors achieved a maximum accuracy of 91.7 with the use of non-linear SVM. The study also translates all the emoticons into text and removed unnecessary punctuation marks. In another study [15], authors used a single layer LSTM model for sentiment analysis with the use of word embeddings and achieve an accuracy of $83.9\%$, in which word embeddings were developed using Skip-gram and CBOW approaches. In [16], authors used LSTM and CNN models on comments data and achieve an accuracy of $65.97\%$ and $54.24\%$ in three and five class labels, respectively. Additionally, the authors used both Skip-gram and CBOW word embeddings.

The study of [17] reports an overview of a shared task on sentiment analysis in Indian languages (SAIL), in which all the participants developed a dataset for sentiment analysis using tweets. In the SAIL 2015, authors in [18] used the Naive Bayes model to classify the SAIL dataset and achieved an accuracy of $33.6\%$. In another study [19], the authors used an SVM-based approach and using the SAIL dataset it reports an accuracy of $42.20\%$. One of the recent studies using the SAIL dataset has been done by Kumar et. al. [20], where the authors proposed dynamic model-based features with a random mapping approach for sentiment analysis, in which authors achieved an accuracy of $95.36\%$.

Most recent works on sentiment analysis are using deep learning techniques. Among them, notable research work includes Long Short-Term Memory (LSTM) [21]–[23], CNN [22], and variational auto-encoder [24]. In [21], authors used SVM, Naive Bayes, and LSTM methods in order to classify the movie reviews.

Our work differs from previous works in a way that we provide a comparative analysis with multiple datasets and diverse classical and deep learning algorithms including transformer models - such as BERT.

## III. Data

In this section, we first discuss the available datasets we used in the study. Following, we analyzed the data to understand how distinctive these lexical content are with respect to the positive and negative sentiment classes.

### A. Data Description

The datasets that we used in our study are:
- **Sentiment Analysis in Indian Languages (SAIL) Dataset** [17]: This dataset, consists of tweet posts, has developed in the Shared task on Sentiment Analysis in Indian Languages (SAIL) 2015. The training, development, and test set of this dataset has 1000, 500, and 500 tweets, respectively. In our study, we only use the training set and split it into train, development, and test set.
- **ABSA Dataset** [25]: This dataset has developed to perform aspect-based sentiment analysis task in Bangla. The dataset contains two categories of data which are cricket and restaurant. In the cricket category, authors collected data from Facebook, BBC Bangla, and Prothom Alo and manually annotated them, and in the restaurant category, authors directly translated the English benchmark's Restaurant dataset [26].
- **BengFastText Dataset** [27]: This data was collected from several newspapers, TV news, books, blogs, and social media. The original dataset reports 320,000 data for sentiment analysis, however, a fraction of it is publicly available. The public version includes 8,420 posts including a test set. We combined the training and test set first, then we split the data into train, development, and test sets.
- **YouTube Comments Dataset** [16]: This dataset was developed by extracting comments from various YouTube videos. The dataset were labeled using three class, five class, and emotion labels. In our study, we only took the data of three class label and converted five class into three class label. As a result, we got total 2,796 comments which we split into train, development, and test sets.
- **Social Media Posts (CogniSenti Dataset)**: This is our in-house developed dataset consists of posts from Twitter and Facebook. We manually annotated the collected posts by native Bangla speakers in three class labels. We have annotated total 6,570 posts in which 942 posts are facebook statuses and 5,628 are tweets. We split this data in train, development, and test set, which contains 4599, 985, and 986 posts, respectively.

In Table I, we report the distribution of the datasets we curated and used in our experiments. For this study, we removed romanized Bangla annotated posts[2] to have a clear view about Bangla content. Hence, the prepared datasets are purely on Bangla.

### B. Data Analysis

One of the major focus of this study was to understand the characteristics of the datasets. For this, we first discussed basic

---
[2]For some datasets such as Youtube comments for example consisted romanized comments.

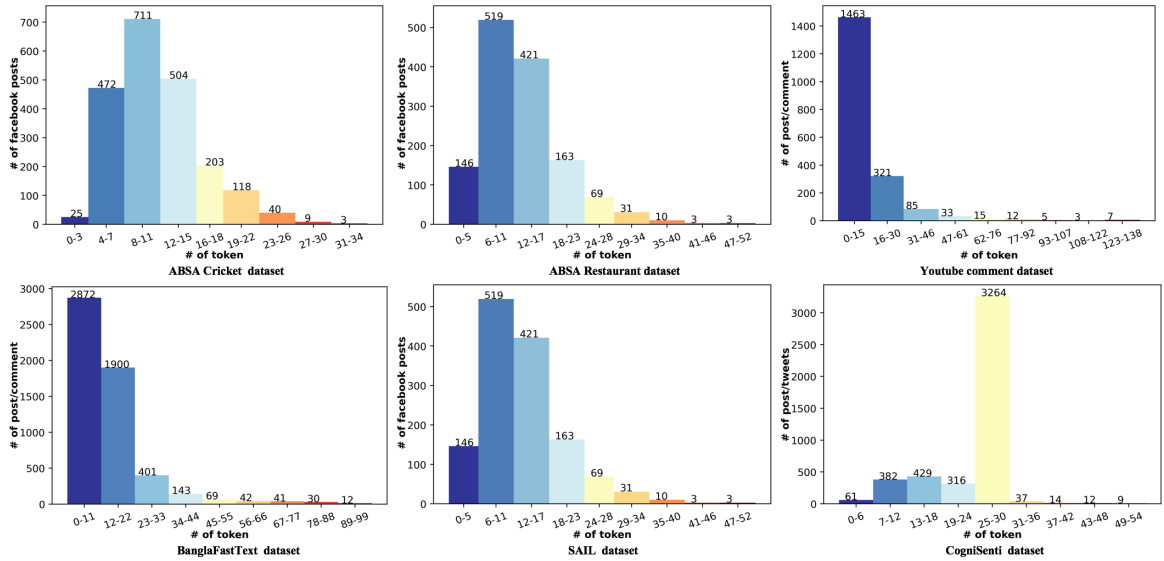

Figure 1: Number tokens in different bins for different datasets

Table I: Data splits and distributions of different datasets

| Class label | Train | Dev | Test | Total |
|---|---|---|---|---|
| **ABSA: Cricket Dataset** | | | | |
| Positive | 376 | 71 | 73 | 520 |
| Neutral | 194 | 27 | 34 | 255 |
| Negative | 1515 | 274 | 273 | 2062 |
| **Total** | 2085 | 372 | 380 | 2837 |
| **ABSA: Restaurant Dataset** | | | | |
| Positive | 872 | 143 | 116 | 1131 |
| Neutral | 167 | 35 | 46 | 248 |
| Negative | 326 | 46 | 57 | 429 |
| **Total** | 1365 | 224 | 219 | 1808 |
| **BengFastText Dataset** | | | | |
| Positive | 2403 | 595 | 788 | 3786 |
| Negative | 3107 | 783 | 744 | 4634 |
| **Total** | 5510 | 1378 | 1532 | 8420 |
| **SAIL** | | | | |
| Positive | 331 | 69 | 71 | 471 |
| Neutral | 194 | 39 | 44 | 277 |
| Negative | 174 | 41 | 35 | 250 |
| **Total** | 699 | 149 | 150 | 998 |
| **Youtube Comments Dataset** | | | | |
| Positive | 553 | 103 | 96 | 752 |
| Neutral | 539 | 106 | 116 | 761 |
| Negative | 865 | 210 | 208 | 1283 |
| **Total** | 1957 | 419 | 420 | 2796 |
| **Social Media Posts (CogniSenti Dataset)** | | | | |
| Positive | 1047 | 205 | 236 | 1488 |
| Neutral | 2633 | 553 | 563 | 3749 |
| Negative | 919 | 227 | 187 | 1333 |
| **Total** | 4599 | 985 | 986 | 6570 |

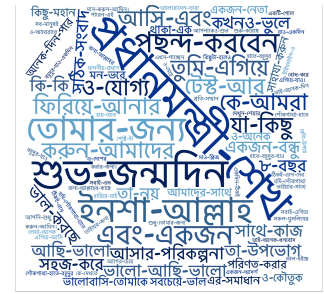

Figure 2: Word-cloud for **highly positive tokens (selected from positive class)**, in the combined datasets with valence score, $\vartheta(.) = 1$. – is used to indicate space between the bi-grams

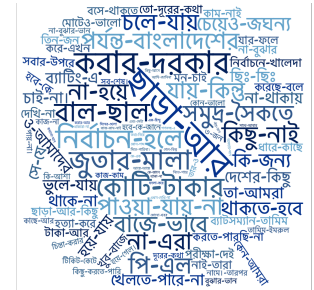

Figure 3: Word-cloud for **highly negative tokens (selected from negative class)**, in the combined datasets with valence score, $\vartheta(.) = 1$. – is used to indicate space between the bi-grams

characteristics of the datasets in terms of token length, and then we analyzed the content of the data for highly significant positive/negative lexical n-grams in individual datasets and in combination.

*a) Token Length Distribution:* The computed distribution of tokens is reported in Figure 1. Across different datasets, the number of tokens is less than 90. The average number of token in the combined dataset is 16 and the median is 13. Only in a very few cases, it is above this range (e.g., Youtube comments, BanglaFastText, and CogniSenti dataset). This analysis helps us in understanding what is the typical maximum number of tokens in social media data that could be used as sequence length – an important hyperparameter for deep learning algorithms.

*b) Class-wise Lexical n-gram Discrimination:* For a better understanding of all sentiment datasets used in this paper and their discrimination power with respect to the categories – positive *vs* negative, we analyzed the lexical content of each dataset individually and in combination. We compared the vocabularies of all classes using the valence score [28], [29], $\vartheta$ for every token, $x$, following the Equation 1:

$$\vartheta(x, L_i) = 2 * \frac{\frac{C(x|L_i)}{T_{L_i}}}{\sum_l^L C(x|L_l)} - 1 \quad (1)$$

where $C(.)$ is the frequency of the token $x$ for a given class $L_i$. $T_{L_i}$ is the total number of tokens present in the class. The $\vartheta(x) \in [-1, +1]$, with $+1$ indicating that the use of the token is significantly higher in that class content than the rest.

Table II: Examples of n-grams in different datasets appeared with postive and negative sentiment posts.

| | ABSA cricket | ABSA restaurant | CogniSenti | BanglaFasttext | SAIL | Youtube |
|---|---|---|---|---|---|---|
| **Positive** | বয়স পর্যন্ত | বন্ধুত্বপূর্ণ সেবা | শুভ জন্মদিন | একটি দুর্দান্ত | আরো জোরদার হবে | সুন্দর নাটক |
| | বাংলাদেশ জিতবে | খাবার এবং পরিবেশ | সরাসরি লাইভ | হাজার বছর | জোরদার হবে প্রধানমন্ত্রী | ভালো লাগে। |
| | ভালো মানের | গ্রেট খাবার | এখনকার প্রধান খবর | তোমার জন্য | মজার মুহূর্ত | ধন্যবাদ আপনাকে |
| | করার পর | সুম্বাদু খাবার | পাঠিয়ে দিন বিবিসি | সাশ্রয়ী মূল্যের | ব্যবস্থায় বিচারকদের স্বাধীনতা | অসাধারণ একটা |
| | বুঝেন ৪০ পেরোনো | এবং চমৎকার | পাতায় দেখা যাবে | সত্যিই ভাল | দর্শকদের হতাশ করবেননা | সত্যিই অসাধারণ |
| | জন্য শুভকামনা | খুব ভাল এবং | সংবাদ সম্মেলনে | হেফাজত কর | অভিযোগে মুখ খুললেন | উপস্থাপক কে ধন্যবাদ। |
| | মানের ব্যাটসম্যান | ভাল মূল্য | আমাদের ইউটিউব | খুব সুন্দর লাগছে | দোয়া করে | এতো সুন্দর হয় |
| | কি সুন্দর খেলে | খাঁটি খাবার | পাবেন লাইভ | বন্ধুত্বপূর্ণ সেবা | শুরু হোক | ভাল লাগল স্যার |
| | এখনো ক্ষুরধার তাহলে | মনোযোগী এবং | যা থাকছে বাংলাদেশে | ক্রিকেটের সকল | একের পর এক | শুভ নববর্ষ |
| | আবার সুযোগ দেয়ার | ভাল সেবা। | যোগাযোগ করবো। | আল্লাহ আমার রব | দূষণ রোধ | ভাল লাগার |
| **Negative** | করা হোক | প্রয়োজন নেই | অশ্লীল ভিডিও | মানুষ না | সোনালী ও প্রাইম | জুতা মারি |
| | দেখে মনে | ছিল অনুপযুক্ত | মাদক ও | দল থেকে বাদ | ইয়াবা ও | তুর গালে |
| | একটা ম্যাচ | অবিচক্ষণ সেবা | ইয়াবা ও | বাদ দেওয়া | অশ্লীল ভিডিও | মাগির পোলা |
| | ক্রিকেট কে | অহংকারী কর্মচারী | কর্মকর্তার বিরুদ্ধে | ছাড়া আর | কর্মকর্তার বিরুদ্ধে | খানিকটা বাংলাদেশের কলঙ্ক |
| | শেষ করে | অনুপযুক্ত এবং | মাদক ও দেহ | গুলি করে | বাল্য বিবাহ | চিলিচিলি করে |
| | দেওয়া হোক | এবং ব্যবস্থাপনার কোনও | কষ্ট নিয়ে | লাভ নেই | চেয়েও জঘন্য | মারার দরকার |
| | করতে পারবে | অযৌক্তিক ছিল | হোটেলে মাদক ও | ঘেউ ঘেউ | নিন্দা ও | তোরে পোরানো হবে।মোল্লা |
| | দল থেকে বাদ | পরিবেশ খারাপ | কপাল খারাপ | সন্ত্রর না | হত্যা সিদ্ধিরগঞ্জে থানার | একটা পাগল |
| | বাদ দেওয়া | ভয়ানক ছিল | আবাসিক হোটেলে মাদক | হত্যা করা | বিরুদ্ধে মামলা | মরলে মন্দিরে তোরে |
| | বুঝি না | জন্য ব্যয়বহুল | বাল্য বিবাহ | মিথ্যা কথা | গুলি উদ্ধা | |

Table II presents top frequent bi- and tri-grams with $\vartheta = 1.0$ for each individual dataset. In addition, we selected tokens present in most of the data that has $\vartheta = 1.0$ per class label. From the table, we observe these n-grams clearly represent the positive and negative aspects of the sentiment. Using these class-specific n-grams a sentiment lexicon can be developed. We made this lists[1] publicly available to enrich the current state-of-art. In addition to calculating the valence, $\vartheta$, for individual dataset, we also extracted a common set of n-grams present in most of the dataset, representing their usage in cross-platform and in multiple domains. Figure 2-3 presents the word-cloud designed with such a common subset. From the figures, positive and negative words and their semantic differences are quite clear.[3]

## IV. Experimental Setup

In this section, we describe the details of our classification experiments and results. To run the experiments, we split data into training, development, and test sets with a proportion of 70%, 15%, and 15%, respectively.

### A. Preprocessing

Social media content is always noisy, which consists of many symbols, emoticons, URLs, username, and invisible characters. Previous studies show that filtering and cleaning the data before training a classifier helps significantly. Hence, we preprocess the data before classification experiments. The preprocessing steps include removal of stop words, invisible characters, punctuations, URLs, and hashtag signs. Note that most of the datasets contain romanized text and we removed them for this study.

---

[3]We are aware that understanding such word-cloud requires native speaker, however, we feel that it is important to report such findings as they clearly reflects positive and negative characteristics.

### B. Models and Architectures

We have conducted classification experiments using both classical and deep learning algorithms. As the classical algorithms, we used the two most popular algorithms *(i)* Random Forest (RF) [30], and *(ii)* Support Vector Machines (SVM) [31].

As deep learning algorithms, we used CNN [32], Fast-Text [33] and transformers based models such as BERT [7], DistilBERT [34] and XLM-RoBERTa [35]. The choice of algorithms is motivated by several factors. CNN has been widely use with static embedding (i.e., word2vec) for many NLP tasks in the past years where FastText extends static embedding to use n-grams (i.e., context). Static embedding based approaches generate a single embedding for each word where transformers based models take contextual information into account and generates contextual embedding. It uses Word-Piece tokenization [36], that is a data-driven approach, which addresses out-of-vocabulary problem. Among the transformer models BERT and XLM-RoBERTa are larger in parameter size (168 million parameters in $BERT_{base\_multilingual}$ and 550 million in $XLM\text{-}RoBERTa_{large}$) whereas DistilBERT consists of 134 million parameters. Network size and the number of parameters also reflect the performance and computation time, which we wanted to understand.

For all models, we used the multilinguial version of the models. We follow the fine-tuning procedure using a task-specific layer on top of the transformers network.

### C. Evaluation

We computed weighted average precision (P), recall (R), F1-measure (F1) to measure the performance of each classifier. We choose the weighted metric, which takes care of the class imbalance problem.

Table III: Classification results using RF, SVM, CNN, FastText, BERT, DistilBERT and RoBERTa. Best results are highlighted with bold form.

| Dataset | RF | SVM | CNN-W2V | CNN-Glove | FastText | BERT | DistilBERT | XLM-RoBERTa |
|---|---|---|---|---|---|---|---|---|
| ABSA cricket | 0.662 | 0.636 | 0.679 | **0.696** | 0.688 | 0.682 | **0.699** | 0.682 |
| ABSA restaurant | 0.407 | 0.498 | 0.391 | 0.491 | 0.519 | 0.581 | 0.579 | **0.692** |
| SAIL | 0.546 | 0.552 | 0.557 | **0.595** | 0.532 | 0.566 | 0.570 | 0.566 |
| BengFastText | 0.612 | 0.613 | 0.663 | 0.657 | 0.661 | 0.674 | 0.669 | **0.674** |
| Youtube comments | 0.586 | 0.605 | 0.669 | 0.663 | 0.658 | **0.729** | 0.701 | **0.729** |
| CogniSenti | 0.545 | 0.584 | 0.604 | 0.587 | 0.614 | **0.686** | 0.589 | **0.686** |
| **Avg.** | **0.560** | **0.581** | **0.594** | **0.615** | **0.612** | **0.653** | **0.635** | **0.671** |

## D. Experiments

*a) Classical Algorithms:* To train the classifiers using the classical algorithms mentioned above, we first converted the preprocessed data into bag-of-$n$-gram vectors weighted with tf-idf. We used $n$-grams (i.e., unigram, bigram, and tri-gram) to utilize contextual information. For both SVM and RF we use grid search to optimize the parameters.

*b) CNN:* To train the model with CNN we used a filter size of 300 filters with a window size of 2, 3, and 4 with the same pooling length. We used Adam optimizer [37] and the maximum number of epochs was set to 3,000. To reduce the computation without impacting the performance we used *early stopping* based on the accuracy on the development set. The patience size in early stopping was set to 200. These parameter settings varied in different experiments. To train the CNN, we experiment with two different in-house developed pre-trained static embedding models i) word2vec skip-gram [38], ii) glove [39]. For training the pre-trained word embedding models we used the dataset reported in [40] along with the latest Bangla Wikipedia dump.

*c) FastText:* For the FastText, we use pre-trained embeddings, which is released by FastText[4] for Bangla. To tune the hyperparameter we use the built-in hyperparameter setups from the FastText.

*d) Transformers models:* For transformer-based models, we used the Transformer Toolkit [41]. We fine-tune each model using a learning rate of $1e-5$ for ten epochs [7]. The training of the pre-trained models has some instability as reported in [7], therefore, we run each experiment 10 times using different random seeds and select the model that performs the best on the development set. With transformer models, we used model specific tokenizer available with Transformer Toolkit.

## V. Results and Discussion

### A. Results

In Table III, we report the classification results for each dataset. It summarizes the results obtained by different classification algorithms described above. Though we computed several metrics for the performance measure, because of the limited space, we only report F1-score. We reported other results in a public repository[1] along with the data split. From the table, we observe that among the classical algorithms the performance of SVM is better than RF except for ABSA cricket dataset where RF is 2.6% higher. As expected we have consistent improvement with deep learning models. Both skip-gram (word2vec) and glove embedding with CNN are performing comparatively. The performance of FastText is higher in three datasets and low on other datasets though overall FastText is better than CNN. Across different results transformer models consistently perform better than others, which highlights the capability of the pretrained transformer models. Among transformer based models, BERT is performing better on three datasets and XLM-RoBERTa is performing well on two datasets.

[4] http://fasttext.cc/

### B. Discussion and Future Works

Our findings on data analysis suggest that computing the valance scores, $\vartheta$, can help to develop a sentiment lexicon in an automated fashion, with a meaningful representation per class. In terms of performance difference between classical vs deep learning algorithms, there is a significant gap. Hence a compromise has to make between computational cost *vs* accuracy to use them in real-time applications. For example, we used the 4 NVIDIA Tesla V100-SXM2-32GB GPU machine consists of 56 cores and 256GB CPU memory for the experiments. To perform a fine-tuning experiment on a single GPU machine, on average it took 40 minutes using a BERT base model. Whereas, the training an SVM model took less than 5 minutes. Note that running/execution time varies depending on the dataset size as well.

From the perspective of performance comparison across datasets, an end to end comparison might not possible because of the different test set, however, from the results, we conclude that we obtain higher weighted F1 on the *Youtube comments* dataset. Among the datasets SAIL is the smallest in size, this could be a reason that we have a low performance on this dataset.

As future work, we will investigate cross model performance, and consolidate the datasets to see if it enhance the performance of the models. Another future study could be aspect-based sentiment while understanding the diversity of the dataset.

## VI. Conclusions

In this study, we have conducted comparative experiments using different annotated sentiment datasets consisting of Bangla content from social media for multiple domains. We curated, cleaned, and split the dataset for the classification experiments. We analyzed the datasets to understand the discriminating aspects of lexical content for positive and negative sentiment. Our analysis, using the valance score, show a reasonable, meaningful, and clearly visible distinction between the classes. We investigated diverse machine learning algorithms ranging from classical (e.g., SVM) to deep learning algorithms (e.g., CNN, transformer models). The performance of deep learning algorithms is comparatively higher compared to classical algorithms. This improvement comes with the cost of increased resource and time complexity. Given that the official splits in most of the datasets are not available, hence, we are not able to compare our results with previous results. With that in mind we release the data split used in this study, for the purpose of reproducibilty and using the obtained results as benchmarks.